\documentclass[10pt,twocolumn,letterpaper]{article}

\usepackage{cvpr}              

\usepackage{graphicx}
\usepackage{amsmath}
\usepackage{amssymb}
\usepackage{booktabs}
\usepackage{mathrsfs}
\newcommand{\co}{\cos(\theta)}
\newcommand{\si}{\sin(\theta)}
\newcommand{\se}{\mathrm{SE}(2)}

\newcommand{\see}{\mathrm{SE}(3)}
\newcommand{\g}{\mathfrak{g}}
\newcommand{\G}{\mathfrak{G}}
\newcommand{\R}{\mathsf{R}}
\newcommand{\Rp}{\mathsf{R}^{(2)}}
\newcommand{\RpT}{\mathsf{R}^{(2)\top}}
\newcommand{\T}{\mathbf{t}}
\newcommand{\Tp}{\mathbf{t}^{(2)}}

\newtheorem{definition}{Definition}

\usepackage[pagebackref,breaklinks,colorlinks]{hyperref}

\usepackage[capitalize]{cleveref}
\crefname{section}{Sec.}{Secs.}
\Crefname{section}{Section}{Sections}
\Crefname{table}{Table}{Tables}
\crefname{table}{Tab.}{Tabs.}

\def\cvprPaperID{10667} 
\def\confName{CVPR}
\def\confYear{2022}

\begin{document}


\title{Leveraging Equivariant Features for Absolute Pose Regression}

\author{Mohamed Adel Musallam\\
{\tt\small mohamed.ali@uni.lu}
\and Vincent Gaudilli\`ere\\
{\tt\small vincent.gaudilliere@uni.lu}
\and Miguel Ortiz del Castillo\\
{\tt\small miguel.ortizdelcastillo@uni.lu}
\and Kassem Al Ismaeil\\
{\tt\small kassem.alismaeil@gmail.com}
\and Djamila Aouada\\
{\tt\small djamila.aouada@uni.lu}
\and
Interdisciplinary Center for Security, Reliability and Trust (SnT)\\University of Luxembourg, Luxembourg
}
\maketitle

\begin{abstract}
While end-to-end approaches have achieved state-of-the-art performance in many perception tasks, they are not yet able to compete with 3D geometry-based methods in pose estimation. Moreover, absolute pose regression has been shown to be more related to image retrieval. As a result, we hypothesize that the statistical features learned by classical Convolutional Neural Networks do not carry enough geometric information to reliably solve this inherently geometric task. In this paper, we demonstrate how a translation and rotation equivariant Convolutional Neural Network directly induces representations of camera motions into the feature space. We then show that this geometric property allows for implicitly augmenting the training data under a whole group of image plane-preserving transformations. Therefore, we argue that directly learning equivariant features is preferable than learning data-intensive intermediate representations. Comprehensive experimental validation demonstrates that our lightweight model outperforms existing ones on standard datasets.\footnote{This work was funded  by  the Luxembourg National  Research  Fund (FNR), under the project reference BRIDGES2020/IS/14755859/MEET-A/Aouada, and by LMO (https://www.lmo.space).}
\end{abstract}

\section{Introduction}
\label{sec:intro}

\begin{figure}[t]
\begin{center}
\includegraphics[width=\linewidth,trim=0cm 1.4cm 0cm 0cm, clip]{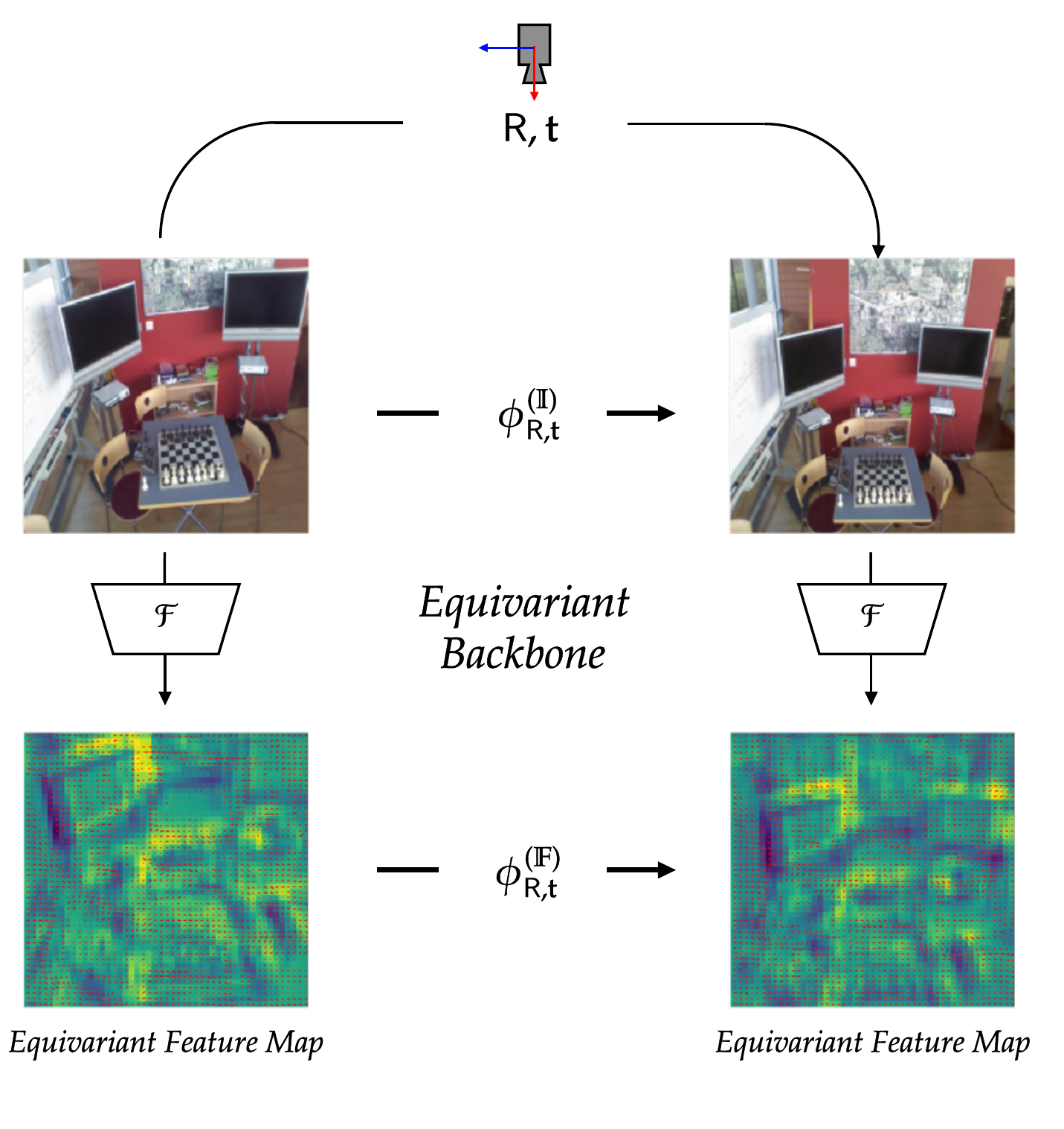}
\end{center}
\caption{
\textbf{Illustration of our approach} - Our method adopts a translation and rotation-equivariant convolutional neural network to extract geometry-aware features that directly encode camera planar motions $\mathsf{R},\mathbf{t}$. While camera moves, equivariance of the proposed feature extractor $\mathcal{F}$ leads to explicit image ($\phi^{(\mathbb{I})}_{\mathsf{R},\mathbf{t}}$) and feature ($\phi^{(\mathbb{F})}_{\mathsf{R},\mathbf{t}}$) changes. This property is leveraged to propose a more efficient solution to the absolute pose regression problem.
}
\label{fig:teaser}
\end{figure}

In computer vision, camera pose estimation, and its reference frame inverse, \textit{i.e.}, object pose estimation, have been extensively studied over the last decades~\cite{SONG2021103055,LepetitF05,MarchandUS16}.
 
Traditionally, pose estimation has been addressed using 3D geometry. In practice, a set of 2D-3D feature correspondences is generated, then statistically leveraged to recover the camera pose~\cite{RANSAC,EPnP,PnL,Pix2Pose}.
More recently, direct Absolute Pose Regression (APR) approaches have been introduced, drawing upon early successes of deep learning~\cite{DNN-survey}. These methods consist in directly mapping an image to its pose using a suitably trained Convolutional Neural Network (CNN). Therefore, end-to-end trainable methods have the advantage of providing fully differentiable results, enabling the optimization of all parameters in a comprehensive manner. Moreover, predictions are achieved at a steady speed and power consumption, whereas RANdom SAmple Consensus (RANSAC)-based methods~\cite{RANSAC} are less predictable and likely to suffer from an efficiency drop when the inlier rate is low. However, state-of-the-art APR methods have been proven theoretically and shown experimentally  to have a lower accuracy compared to 3D structure-based approaches~\cite{Sattler_2019_CVPR}. Indeed, the former are more closely related to image retrieval than to 3D structure~\cite{Sattler_2019_CVPR}.

The questions we ask in this work are: \emph{Why do current APR methods fall short in accuracy ? How can they reach their full potential?} 
Our hypothesis is that there is a lack of exploitation of the geometric properties of data. This happens typically at the level of the feature extraction layers commonly used in classical deep learning approaches. 
Specifically, we posit that in the case of APR and pose estimation, having a representation which is \emph{equivariant} to the group of rigid motions, \textit{i.e.,} rotations and translations in 3D, may be an effective way to boost the network performance. This should play the role of an implicit data augmentation by means of group equivariance, and in turn alleviate the need for an explicit data augmentation for training. \\
Indeed, recently, there has been a growing interest in designing more geometric models that are equivariant to groups of such transformations. These approaches leverage theoretical contributions from group theory, representation theory, harmonic analysis and fundamental deep learning~\cite{cohen2016group,cohen2017steerable,cohen2018spherical,gen-E2-eq,Chen_2021_CVPR,Han_2021_CVPR,Gupta_2021_CVPR}. More specifically, group-equivariant neural networks, or Group-equivariant CNNs (G-CNNs), are part of the broader and promising field of geometric deep learning~\cite{gdl_book}, that aims to exploit any underlying geometric relationship that can exist within the data. In particular, the special Euclidean groups in 2 and 3 dimensions, denoted as $\se$ and $\see$ and encompassing respective rigid motions, are of particular interest in 3D computer vision~\cite{gen-E2-eq,EstevesSLDM19}.

Despite the conceptual advances they represent, to the best of our knowledge, the use of deep equivariant features in the APR context is still considerably unexplored.
This paper proposes, for the first time, to investigate and justify the use of deep equivariant features for solving APR (see Figure \ref{fig:teaser}).

\vspace{1mm}
\noindent\textbf{Contributions.}
Our contributions are summarized below:\\
(1) A formulation of how an equivariant CNN induces representations of planar camera motions, lying in $\se$, directly into the feature space. (Section \ref{ssec:theory-1})\\
(2) An intuitive explanation is provided as to how $\se$-equivariant features can be leveraged to recover any camera motion lying in $\see$. (Section \ref{ssec:theory-2})\\
(3) A lightweight equivariant pose regression model, referred to as \textit{E-PoseNet}, is introduced. (Section \ref{sec:method})\\
(4) Extensive experimental evaluation of E-PoseNet showing its competitive performance as compared to existing APR methods on standard datasets. (Section \ref{sec:experiments})

\vspace{1mm}
\noindent\textbf{Paper organization.}
An overview of state-of-the-art APR methods and current exploitations of deep equivariant features is given in Section~\ref{sec:rel-work}. Section~\ref{sec:prelim} presents the formal definition of equivariance along with the formulation of APR. The theoretical justification as to how $\se$-equivariant features can explicitly encode planar camera motions is presented in Section~\ref{sec:theory}, whereas the full pose regression pipeline is introduced in Section~\ref{sec:method}. An extensive experimental validation is given in Section~\ref{sec:experiments} along with a discussion of limitations. Section~\ref{sec:conclusions} concludes the paper.

\section{Related Work}
\label{sec:rel-work}
The goal of this paper is to exploit the power of equivariant features in the context of APR. Therefore, we split this section into: (1) a review of the relevant literature on APR, and (2) an overview of recent deep equivariant feature extraction methods applied to computer vision problems.

\vspace{1mm}
\noindent\textbf{Absolute Pose Regression.}
Since the rise of deep learning and CNNs in the early 2010's, many works have explored the application of CNNs for APR. This began with the introduction of PoseNet by Kendal \textit{et al.}~\cite{PoseNet}, who used the GoogLeNet model~\cite{GoogLeNet} as a feature extraction backbone coupled with a regression head to estimate the translation and rotation vectors.
Most of the subsequent improvements lie in changes in the feature extraction architecture~\cite{PoseNet,apr-icra-2017,brahmbhatt2018geometry}, modified objective functions~\cite{KendallC16,WalchHLSHC17,MelekhovYKR17}, and additional intermediate representations~\cite{Hu_2019_CVPR,Hodan_2020_CVPR}.

In~\cite{Sattler_2019_CVPR}, Sattler \textit{et al.} provide an in-depth analysis of existing works on APR~\cite{apr-icra-2017,apr-iros-2017,apr-cvpr-2017,apr-bmvc-2018,apr-rss-2018}. In particular, they show that structure-based and image retrieval methods are more accurate than APR. Moreover, they demonstrate that APR algorithms do not explicitly leverage knowledge about projective geometry. Instead, they learn a mapping between image content and camera poses directly from the data, and in the form of a set of base poses such that all training samples can be expressed as a linear combination of these reference entities. Wang \textit{et al.}~\cite{Wang_2021_CVPR} proposed an approach to integrate dense correspondence-based intermediate geometric representations within an end-to-end trainable pipeline. However, this method still relies on classical (non-equivariant) features, and thereby requires a significant amount of data for generalization. Furthermore, methods such as~\cite{Hu_2019_CVPR,Hodan_2020_CVPR,Wang_2021_CVPR} propose to learn intermediate representations that are indirectly equivariant, such as segmentation masks, object detections, and depth or normal maps. However, this comes at the cost of parameter redundancy. 
This core observation suggests that directly learning equivariant features may be a valuable direction to improve the accuracy of pose estimation while reducing the number of model parameters. 

\vspace{1mm}
\noindent\textbf{Deep Equivariant Features.} There is a rich history in computer vision on the design of hand-crafted equivariant features (\textit{e.g.,} Scale-Invariant Feature Transform (SIFT)~\cite{SIFT}, Oriented filters~\cite{YokonoP04}, Steerable filters~\cite{FreemanA91}, Rotation-equivariant Fields of Experts (R-FoE)~\cite{Schmidt_2012_CVPR}, Lie groups-based filters~\cite{ferraro1994lie,NordbergG96}). In the deep learning literature, convolutional layers~\cite{CunBDHHHJ89} have been proven to be equivariant to image shifting, while max-pooling layers are only invariant to small shifts of the input image~\cite{Goodfellow-et-al-2016}.\\
Although convolutional layers are inherently equivariant to translation, there is a significant amount of spatial information regarding the inputs that is not encoded by CNNs in a precise fashion~\cite{Kayhan_2020_CVPR,islam2021position}. More specifically, local and global poolings, if added to CNNs, render translation information unrecoverable, discarding the foregoing equivariance~\cite{liao2017exact}.\\
A recent investigation shows that many neurons in CNNs learn slightly transformed (\textit{e.g.,} rotated) versions of the same basic feature~\cite{olah2020naturally}. These are especially common in early vision, \textit{e.g.,} in curve detectors, high-low frequency detectors, and line detectors.

There have been attempts to extend the G-CNNs to wider groups of transformations. In~\cite{BrunaM13,Oyallon_2015_CVPR}, Mallat \textit{et al.} extended CNNs to be equivariant to $\se$ using scattering transform with predefined wavelets. In~\cite{BekkersDBR14,JanssenJBBD18}, Bekkers \textit{et al.} also extended CNNs to be equivariant to the $\se$ group via B-splines. In~\cite{cohen2016group}, Cohen \textit{et al.} proposed group convolutions network equivariant to the p4m discrete group via $90^{\circ}$ rotations and flips, where they demonstrated the effectiveness of group convolutions for classification task.\\
More recently, the use of equivariant features has been investigated for solving various computer vision tasks such as 3D point cloud analysis~\cite{Chen_2021_CVPR}, aerial object detection~\cite{Han_2021_CVPR} and 2D tracking~\cite{Gupta_2021_CVPR}.
In \cite{esteves2019cross}, Esteves \textit{et al.} proposed to use projection and embedding from 2D images into a spherical CNN latent space to estimate the relative orientations of the object. Similarly, Zhang \textit{et al.} proposed to use spherical CNN for learning camera pose estimation in omnidirectional localization~\cite{Zhang_2020_ACCV}.
However, to the best of our knowledge, equivariant features have not yet been explicitly leveraged in the context of APR for single 2D input image, which is the very focus of this paper.

\section{Preliminaries}
\label{sec:prelim}
This section provides the necessary mathematical background. First, we introduce the notions of invariant and equivariant features. Then, we present the general framework for APR, and finally show the added value of relying on equivariant features in this context.

\vspace{1mm}
\noindent\textbf{Notation.}
The following notation will be adopted: vectors and column images are denoted by boldface lowercase letters $\mathbf{x}$, matrices by uppercase letters $\mathsf{X}$, scalars by italic letters $x$ or $X$, functions as $\mathcal{X}$ and spaces as $\mathbb{X}$. The special orthogonal group, the Euclidean group, and the special Euclidean group, of dimension $n$, are denoted as $\mathrm{SO}(n)$, $\mathrm{E}(n)$ and $\mathrm{SE}(n)$, respectively.

\vspace{1mm}
\noindent\textbf{Invariant and Equivariant Features.} 

Given an image $\mathbf{x}$, captured by a camera, an APR method $\mathcal{P}$ predicts the 6-Degrees-of-Freedom (6-DoF) pose, \textit{i.e.,} position and orientation, of the camera with respect to its environment. \\
Let us denote by $\mathbb{I}\subset\mathbb{R}^m$ the linear space of vectorized $m$-dimensional images (or image regions), and by $\mathbb{F}\subset\mathbb{R}^n$ the latent space of features -- with dimension $n$. Considering a CNN-based feature extraction function $\mathcal{F}$, we write:
\begin{align*}
    \mathcal{F}:\hspace{1.5mm} \mathbb{I} &\rightarrow \mathbb{F}\\
    \mathbf{x}&\mapsto \mathcal{F}(\mathbf{x}).
\end{align*}
Given $\G$, a generic group of transformations and $\g$, an element of $\G$, we denote by
$\phi_\g^{(\mathbb{I})}$ and $\phi_\g^{(\mathbb{F})}$ the actions of $\g$
into the image and feature spaces, respectively. 
\begin{definition}
$\mathcal{F}$ is invariant to $\mathfrak{G}$ \textit{if and only if}
\begin{equation}\label{eq:invariance}
    \forall \mathfrak{g}\in\mathfrak{G}, \forall \mathbf{x}\in \mathbb{I}, \quad \mathcal{F}(\phi_\g^{(\mathbb{I})}\mathbf{x})=\mathcal{F}(\mathbf{x}).
\end{equation}
\end{definition}

\begin{definition}
$\mathcal{F}$ is equivariant to $\mathfrak{G}$ \textit{if and only if}
\begin{equation}\label{eq:equivariance}
     \forall \mathfrak{g}\in\mathfrak{G}, \forall \mathbf{x}\in \mathbb{I}, \quad \mathcal{F}(\phi_{\g}^{(\mathbb{I})}\mathbf{x})=\phi_{\g}^{(\mathbb{F})}\mathcal{F}(\mathbf{x}).
\end{equation}
\end{definition}
Note that invariance can be seen as a special case of equivariance where $\phi_\g^{(\mathbb{F})}=\mathcal{I}$, the identity mapping, $\forall \mathfrak{g}\in\mathfrak{G}$.

\vspace{3mm}
\noindent\textbf{Equivariant Features for APR.}
Sattler \textit{et al.} proposed the following formulation for the pose function $\mathcal{P}$ ~\cite{Sattler_2019_CVPR}:
\begin{equation}\label{eq:APR}
 \mathcal{P}(\mathbf{x}) = \mathbf{b} + \mathsf{P}.\mathcal{E}(\mathcal{F}(\mathbf{x})),
\end{equation}
where the feature extractor $\mathcal{F}$ is first applied to the image $\mathbf{x}$ followed by a non-linear embedding of the features $\mathcal{E}$ lifting them to a higher-dimensional 
space. Then, a linear projection into the space of camera poses, represented by a matrix $\mathsf{P}$, is applied. Finally, a bias term $\mathbf{b}$ is added.\\
As presented in Section~\ref{sec:rel-work}, the work in~\cite{Sattler_2019_CVPR} demonstrated that classical APR is more closely related to pose approximation via image retrieval than to accurate pose estimation leveraging the 3D structure, thus the accuracy gap.

Our hypothesis is that this is likely due the lack of geometric information carried by classical CNN features. Indeed, the perceptive power of classical CNNs can most often be considered as a statistical phenomenon, whereas pose estimation is a geometrical problem.

In this work, we thus propose to replace the classical convolutional layers of the feature extractor $\mathcal{F}$ by their group-equivariant counterparts~\cite{cohen2016group}, then to assess how this affects both the accuracy and data efficiency of the model.\\
Therefore, assuming that $\mathcal{F}$ is equivariant to $\G$, \textit{i.e.}, verifies \textbf{Definition~2}, and applying the transformation $\phi_{\g}^{(\mathbb{I})}$ to the image $\mathbf{x}$, the pose regression function $\mathcal{P}$ in (\ref{eq:APR}) becomes:
\begin{equation}\label{eq:APREquivariant}
 \mathcal{P}(\phi_{\g}^{(\mathbb{I})}\mathbf{x}) = \mathbf{b} + \mathsf{P}\cdot\mathcal{E}\left(\phi_{\g}^{(\mathbb{F})}\mathbf{v }\right),
\end{equation}
where $\mathbf{v}=\mathcal{F}(\mathbf{x})$.
This suggests that any action of $\G$ on the image has a direct effect in the latent space and that, in particular, camera motion transformations of images, \textit{i.e.} changes in camera pose, explicitly induce actions on the feature vector $\mathbf{v}$, and implicitly on the regressed pose.\\
Considering $\G$ as $\se$ or $\see$, we posit that such equivariant features will help improve the performance of APR.

\section{Pose from SE(2)-Equivariant Features}
\label{sec:theory}

We consider a piecewise planar scene, where the scene planes are parallel to the image plane. We then consider camera motions that locally preserve the latter.

\subsection{SE(2)-Equivariant Features}
\label{ssec:theory-1}

\begin{figure}[t]
\begin{center}
\includegraphics[width=\linewidth, trim=0cm 0.5cm 0cm 0cm, clip]{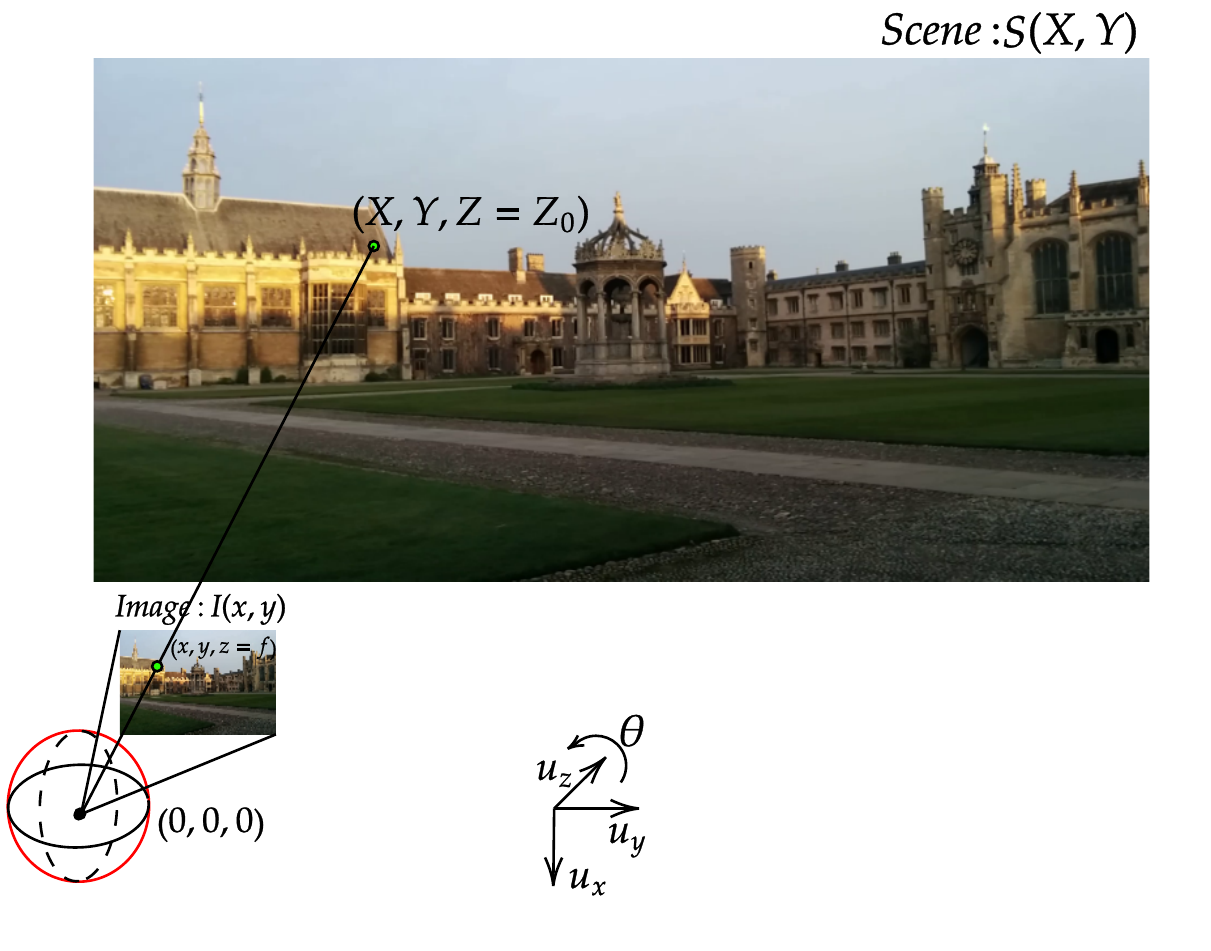}
\end{center}
\caption{
\textbf{Illustration} -- Scene plane $(Z=Z_0)$ - Parallel image plane $(Z=f)$ - Camera center $(0,0,0)$. The scene is defined as the set of light intensity values $\mathsf{S}(X,Y)$ within the plane $Z=Z_0$. Rays of lights are then projected to the camera center. Intersections of projected rays with the image plane (considered as infinite) define image intensity values $\mathsf{I}(x,y)$. Camera motions are restricted to $\se$, \textit{i.e.} translation along $\mathbf{u}_x$, $\mathbf{u}_y$  and rotation around $\mathbf{u}_z$ (characterized by roll angle $\theta$).
}
\label{fig:illustration}
\end{figure}

We herein restrict camera motions to those of the $\se$ group, \textit{i.e.,} planar translations and rotations within the image plane (Figure~\ref{fig:illustration}). With that, we analyse the effects of planar camera motions on the image and feature spaces, assuming that the feature extractor $\mathcal{F}$ is equivariant to $\se$.

\vspace{2mm}
\noindent\textbf{Effects of Camera Planar Motions on Images.}
Following the notations introduced in Figure \ref{fig:illustration}, rotating the camera with a roll angle $\theta$ is equivalent to rotating the scene around $\mathbf{u}_z$ (camera viewing direction) with angle $-\theta$~\cite{Kanatani1990}.\\
Similarly, translating the camera center along $\mathbf{u}_x$ and $\mathbf{u}_{y}$ is equivalent to translating the scene in the opposite direction.\\
Let us denote any rigid motion of the camera along its image plane (\textit{i.e.,} in $\se$) by $\R,\T$, where $\T$ is a planar translation and $\R$ a planar rotation. The effect of this motion on any point $\mathbf{p}$ of the scene is obtained by applying $-\T$ then $\R^{\top}$ such that $\mathbf{p'}=\R^{\top}(\mathbf{p}-\T)$.\\
In 3D, considering $\T=(T_X,T_Y,0)^\top$, we have
\begin{equation}\label{eq:rm_scene}
    \mathbf{p^\prime}=\begin{pmatrix}
    X'\\Y'\\Z'
    \end{pmatrix}
    =\begin{pmatrix}
    \co & \si & 0\\-\si & \co & 0\\0 & 0 & 1
    \end{pmatrix}\begin{pmatrix}
    X-T_X\\Y-T_Y\\Z
    \end{pmatrix}.
\end{equation}
Following classical projection rules~\cite{Hartley2004}, image coordinates $(x,y)$ are then given by $x=f\frac{X}{Z_0}$ and  $y=f\frac{Y}{Z_0}$, where $f$ is the distance from the camera center to the image plane, and $Z_0$ is the distance to the scene plane.\\
Multiplying \eqref{eq:rm_scene} by $\frac{f}{Z_0}$, then restricting the coordinates to the first two ones gives:
\begin{equation}
    \begin{pmatrix}
    x'\\y'
    \end{pmatrix}
    =\begin{pmatrix}
    \co & \si\\-\si & \co
    \end{pmatrix}\begin{pmatrix}
    x-t_x\\y-t_y
    \end{pmatrix},
\end{equation}
where $t_x=f\frac{T_X}{Z_0}$, and $t_y=f\frac{T_Y}{Z_0}$. 
By denoting $\Rp$ the 2D rotation matrix of angle $\theta$ and $\Tp=(t_x,t_y)^\top$, the effect of any planar camera motion $\R,\T$ on any point $\mathbf{p}^{(2)}$ of the image is thus given by
$\mathbf{p}^{(2)}{'}=\RpT(\mathbf{p}^{(2)}-\Tp)$, where
$\mathbf{p}^{(2)}{'}$ is the image of $\mathbf{p}^{(2)}$ under the transformation.\\
We finally denote $\phi^{(\mathbb{I})}_{\R,\T}$ the effect of the camera motion on an image $\mathbf{x}_1$, resulting in another image $\mathbf{x}_2$, \textit{i.e.,} $\mathbf{x}_2=\phi^{(\mathbb{I})}_{\R,\T}\mathbf{x}_1$.\\

\begin{figure}[t]
\begin{center}
\includegraphics[width=\linewidth, trim=0cm 0.2cm 0cm 0cm, clip]{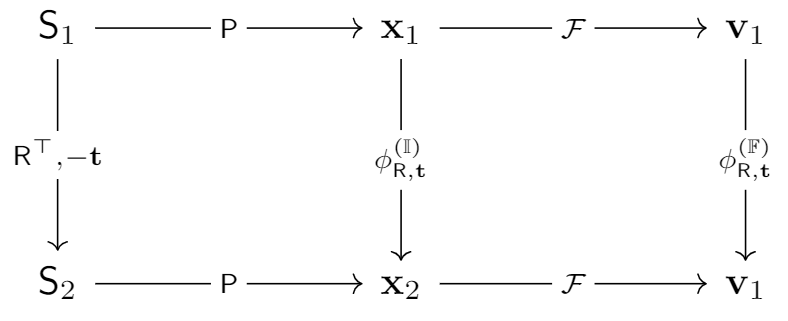}
\end{center}
\caption{
\textbf{Equivariance map -- }
Camera planar motion transformations of planar scenes ($\mathsf{S}_1\mapsto \mathsf{S}_2$, first column), images ($\mathbf{x}_1\mapsto \mathbf{x}_2$, second column) and features ($\mathbf{v}_1\mapsto \mathbf{v}_2$, third column) induce representations of $\se$. In other words, these transformations commute with scene projector $\mathsf{P}$ and feature extractor $\mathcal{F}$.
}
\label{fig:spaces}
\end{figure}
In what follows, we prove that the image transformation due to planar motions of the camera commute with the projection operator $\mathsf{P}$ (Figure \ref{fig:spaces}).
Indeed, applying a planar motion $\R_1,\T_1$ followed by a second one $\R_2,\T_2$ to the camera, 
has the following effect\footnote{Please refer to the supplementary material for further details}: $\mathbf{p'}=\R_3^{\top}\left(\mathbf{p}-\T_3\right)$, where $\R_3=\R_2\R_1$ and $\T_3=\T_1+\R_1\T_2$.
\\
Similarly, one can easily observe that combining two camera motions has a similar effect on any point $\mathbf{p}^{(2)}$ of the image such that $\mathbf{p}^{(2)}{'}=\RpT_3(\mathbf{p}^{(2)}-\Tp_3)$
where $\Rp_3=\Rp_2\Rp_1$, and $\Tp_3=\Tp_1+\Rp_1\Tp_2$. Therefore,
\begin{equation}
    \phi^{(\mathbb{I})}_{(\R_2,\T_2)\circ(\R_1,\T_1)}=\phi^{(\mathbb{I})}_{\R_3,\T_3}=\phi^{(\mathbb{I})}_{\R_2,\T_2}\circ \phi^{(\mathbb{I})}_{\R_1,\T_1}.
\end{equation}
This proves that the correspondence from $\R,\T$ to $\phi^{(\mathbb{I})}_{\R,\T}$ is a group homomorphism from $\se$. In other words, the set of $\phi^{(\mathbb{I})}_{\R,\T}$, where $\R,\T\in\se$, is the image of a representation of $\se$ into the image space.

\begin{figure}[t]
\begin{center}
\includegraphics[width=\linewidth]{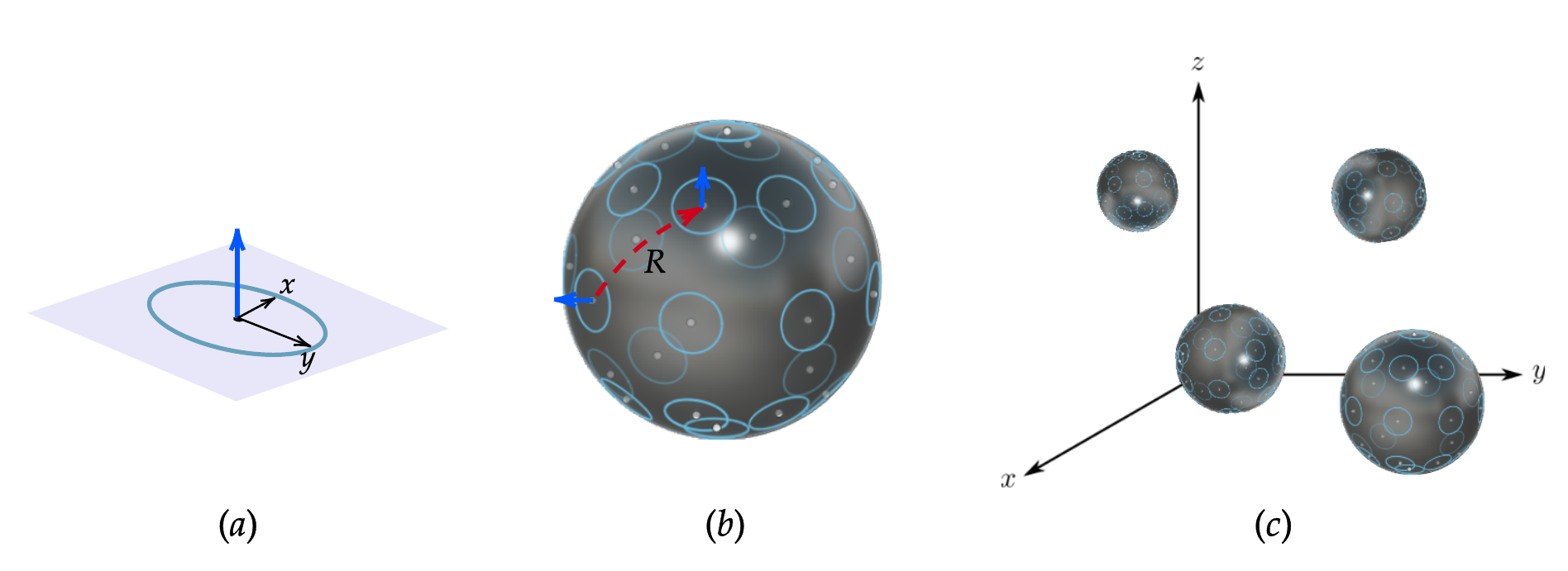}
\end{center}
\caption{
\textbf{From planar to 3D motions -- } (Figure reproduced from~\cite{CohenGW19}): (a) $\mathbb{S}(1)$, (b) $\mbox{SO(3)}/\mbox{SO(2)} \simeq \mathbb{S}(2)$  , (c) $\see = \mathbb{R}^{3} \rtimes \mbox{SO(3)}$.
}
\label{fig:hyperplanes}
\end{figure}

\vspace{1mm}
\noindent\textbf{Effects of Camera Planar Motions on Features.}
We herein consider an $\se$-equivariant CNN-based feature extractor $\mathcal{F}$. For the sake of clarity and simplicity, we discard the discreteness of numerical images and consider their supports as continuous.\\
Classical convolutional layers are only equivariant to the translation group $(\mathbb{R}^{2},+)$. Indeed, at each layer $l$, a conventional CNN takes as input a stack of intermediate feature maps $\mathbf{v}^{(l)}$ : $\mathbb{R}^{2} \rightarrow \mathbb{R}^{K^{(l)}}$ and convolves it with a set of $K^{(l+1)}$ filters $\psi^{(l)}: \mathbb{R}^{2} \rightarrow \mathbb{R}^{K^{(l)}}$. Therefore we have
\begin{equation}
    \forall \Tp\in\mathbb{R}^2,
    \quad
    \left(\left(\phi_{\Tp} \mathbf{v}\right) * \psi^{(l)}\right)(.)=\left(\phi_{\Tp}\left(\mathbf{v} * \psi^{(l)}\right)\right)(.),
\end{equation}
where $\phi_{\Tp}$ are images of $\Tp$ under representations of $(\mathbb{R}^{2},+)$.
In other words, if the input image is translated, the output feature map translates in the same way. However, in general, the same is not true for rotations, \textit{i.e.} if the input image is rotated, the output feature map will not be rotated accordingly. The work in~\cite{gen-E2-eq} has extended CNN equivariance to the $\se$ group, \textit{i.e.} the group of continuous rotations and translations in $\mathbb{R}^{2}$, the image domain.\\
By replacing the translation group equivariance of classical CNNs by equivariance to $\mathrm{SE}(2)$, such particular CNNs can then be characterized by the following equation:
\begin{equation}
\begin{split}
    \forall&\Rp,\Tp\in\se,\\
    &\quad\left(\left(\phi_{\Rp,\Tp} \mathbf{v} \right) * \psi^{(l)}\right)(.)=\left(\phi_{\Rp,\Tp}\left(\mathbf{v} * \psi^{(l)}\right)\right)(.),
\end{split}
\end{equation}
where $\phi_{\Rp,\Tp}$ are images of $\Rp,\Tp$ under representations of $\se$. In particular, considering the last convolutional layer output, we obtain that feature extraction $\mathcal{F}$ and Euclidean transformations of images commute.\\
Finally, the camera motion transformations of both images and features induce representations of $\se$. As a result, the image and feature spaces explicitly encode the planar motions of the camera.

\subsection{From SE(2) to SE(3)}
\label{ssec:theory-2}
After demonstrating how an $\se$-equivariant CNN can induce representations of planar camera motions directly into the feature space, we herein discuss how these features, which are equivariant to planar camera motions, \textit{i.e.,} in $\se$, are leveraged for general pose regression in $\see$. 
\begin{figure*}[t]
\begin{center}
\includegraphics[width=\linewidth,trim=0cm 7.5cm 0cm 3cm ,clip]{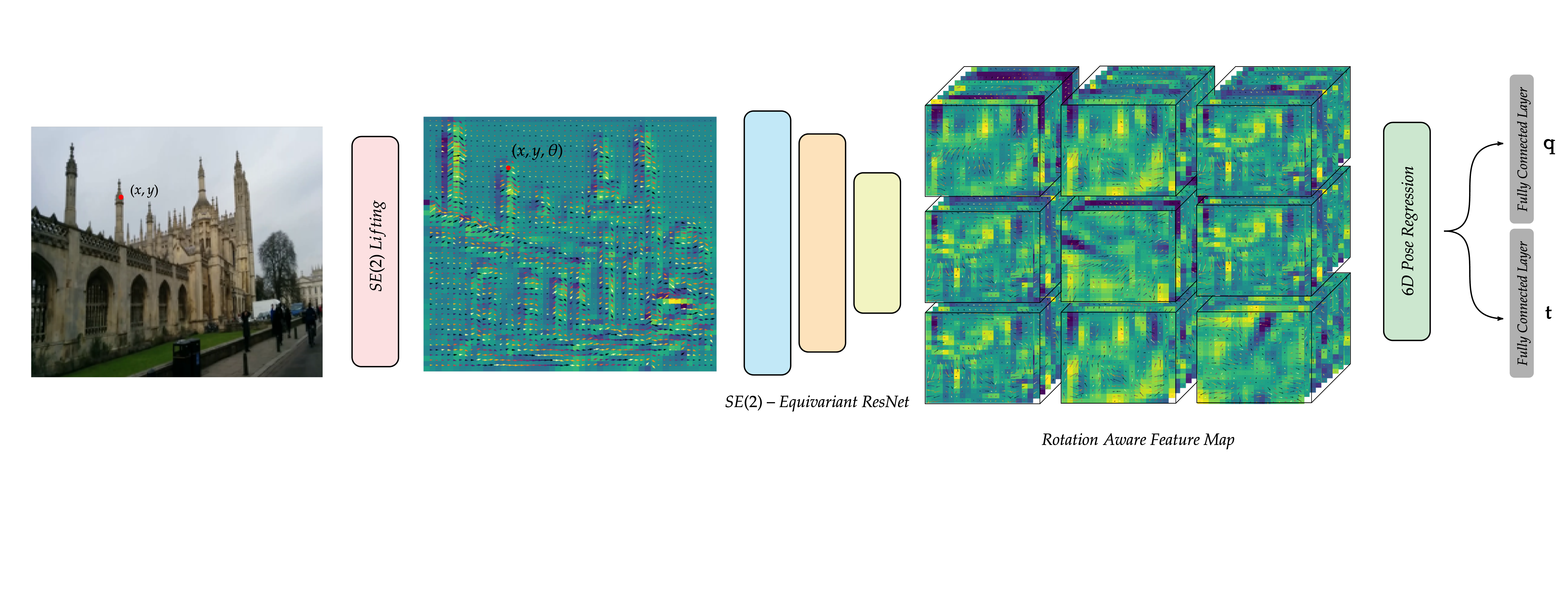}
\end{center}
\caption{
\textbf{E-PoseNet - }
Our pose regression pipeline leverages a roto-translation equivariant ResNet18 \cite{ResNet} backbone, two fully connected Multilayer Perceptrons (MLP) for lifting the features to a higher dimensional space, followed by two branches for separately regressing the position and orientation of the camera.
}
\label{fig:architecture}
\end{figure*}

Indeed, $\se$-equivariant features extracted with $\mathcal{F}$, are now to be mapped to camera poses in $\see$ via $\gamma:=\mathsf{P}\cdot\mathcal{E}$. This is the last step towards finding $\mathcal{P}(\mathbf{x})$, as defined in equation \eqref{eq:APR} \footnote{For the sake of clarity, and without loss of generality, we drop the bias $\textbf{b}$ in this discussion.}.
\\
The $\see$ group may be written as a semidirect product $\see = \mathbb{R}^{3} \rtimes \mbox{SO(3)}$, and similary for $\se = \mathbb{R}^{2} \rtimes \mbox{SO(2)}$. 
We may therefore restrict the discussion to the mapping $\gamma^{*}: \mbox{SO(2)} \rightarrow\mbox{SO(3)}$~\cite{CohenGW19,gdl_book}. We rely on the observation that the quotient space $\mbox{SO(3)}/\mbox{SO(2)} \simeq \mathbb{S}(2)$ is the sphere in 3D, where $\simeq$ denotes a homeomorphism. \\
For every point on $\mathbb{S}(2)$, it is possible to move to another point via a rotation. $\mathbb{S}(2)$  is consequently a homogeneous space for $\mbox{SO}(3)$, and $\mbox{SO}(3)$ can be seen as a bundle of elements of $\mathbb{S}(1)$, \textit{i.e.,} planar circles on $\mathbb{S}(2)$, for which a continuous mapping $\gamma^{*}$ exists~\cite{CohenGW19}. 
These mappings, $\gamma^{*}$, directly relate to $\gamma$ by composing with translations. Figure~\ref{fig:hyperplanes} illustrates how the planar rotations around a fixed axis (Figure~\ref{fig:hyperplanes}(a)) can be viewed as local patches on the sphere in 3D (Figure~\ref{fig:hyperplanes}(b)); thus, relating to rotations in 3D, and finally how this may be generalized to full rigid motions by translation as shown in Figure~\ref{fig:hyperplanes}(c). The mapping $\gamma$
is learned as part of the end-to-end APR.   \\
Intuitively, one can interpret this as an approximation of the space of camera poses by a finite set of learned ones, with feature equivariance used to generalize and extend the coverage within the space. Indeed, an $\se$-equivariant model is capable of generalizing from each learned pose to every poses that preserve the image plane (\textit{i.e.} z-rotated and x,y-translated versions of the original camera). In other words, instead of learning some cropped image planes like classical CNNs do, relying on an $\se$-equivariant CNN rather consists in learning several infinite image planes, therefore providing a denser coverage of the scene space.

\section{Proposed  \textit{E-PoseNet}}
\label{sec:method}

This section gives an overview of our proposed equivariant pose regression model, \emph{E-PoseNet}.\\
To be able to assess how explicitly encoding pose information into the feature space can result in a more accurate and data-efficient pose regressor, we proceed from the architecture of PoseNet~\cite{PoseNet}. We follow the same pipeline, except that we substitute the GoogLeNet backbone by an $\se$-equivariant~\cite{gen-E2-eq} version of ResNet~\cite{ResNet}, to extract both translation and rotation-equivariant features. The resulting model is presented in Figure~\ref{fig:architecture}.

\vspace{1mm}
\noindent\textbf{Network Architecture.}
\textit{E-PoseNet} is composed of a roto-translation equivariant ResNet18 backbone, two fully connected Multilayer Perceptrons (MLP) for lifting the features to a higher dimensional space, followed by two branches for separately regressing the position and orientation of the camera. Each branch consists of an independent fully-connected MLP head.

\vspace{1mm}
\noindent\textbf{SE(2)-Equivariant Backbone.}
Our backbone, \textit{i.e.,} feature extractor $\mathcal{F}$, is an $\se$ roto-translation equivariant version of ResNet. Specifically, we use the \emph{e2cnn}~\cite{gen-E2-eq} implementation for $\mathrm{E}(2)$-equivariant convolution, pooling, normalization, and non linearities, to build an equivariant ResNet18.\\
To decrease the computational cost, we discretize the $\se$ group making the model only equivariant to the $\left(\mathbb{R}^{2},+\right) \rtimes C_{N}$ group, meaning all translations in $\mathbb{R}^{2}$ and rotations by angles multiple of $\frac{2\pi}{N}$.
Extracted features are now rotation-equivariant feature maps $\mathsf{V}$ with the size $( K \times N \times H \times W )$, where $K$ is the number of channels, $N$ the number of feature orientations (for our model we used  $N = 8$), and $H,W$ respectively, the height and width.\\
In addition to classical translation information, obtained features thus encode rotation information that can enhance the pose regression. Furthermore, equivariance to broader transformations constrains the network in a way that can aid generalization, especially due to the weights shared under image rotations \cite{cohen2016group}. Finally, this rotation-equivariant ResNet shows a significant reduction in model size, about $1/N$ parameters compared to the regular ResNet architecture, to obtain the same feature size. Indeed, the size of classical feature maps is in the form $( K \times H \times W )$.

\vspace{1mm}
\noindent\textbf{Loss Function.} To regress camera poses, we use the loss function introduced in~\cite{apr-cvpr-2017}, and defined as:
\begin{equation}
\label{eq:loss}
\mathcal{L_P}=\mathcal{L}_\T \exp \left(-s_{\T}\right)+s_{\T}+\mathcal{L}_{\R} \exp \left(-s_{\R}\right)+s_{\R},
\end{equation}
where the position loss $\mathcal{L}_{\T}=\left\|\T_{0}-\T\right\|_{2},$ and the orientation loss $\mathcal{L}_{\R}=\left\|\mathbf{q}_{0}-\frac{\mathbf{q}}{\left\|\mathbf{q}\right\|_{2}}\right\|_{2},$
are computed from predicted ($\mathbf{q}$,$\mathbf{t}$) and groundtruth ($\mathbf{q}_0$,$\mathbf{t}_0$) camera poses, considering the quaternion representation for orientations. $s_{\T}$, $s_{\R}$ are learned parameters.

\section{Experiments and Analysis}
\label{sec:experiments}

The proposed method aims to improve APR accuracy by utilizing an equivariant feature extraction backbone able to learn geometry-aware feature maps. We first show the effect of $\se$-equivariant models on rotated feature maps using samples from the T-Less dataset~\cite{T-LESS}. Then, we benchmark our proposed \textit{E-PoseNet} on two datasets for both indoor and outdoor camera localization. 

\vspace{1mm}
\noindent\textbf{Equivariance analysis on T-Less.}
In this study, we use a sequence of `object 5' from the T-less training dataset~\cite{T-LESS}. With only one textureless symmetric object present in the scene and undergoing continuous rotations, this sequence represents an ideal case for testing the impact of the different rotation parametrization and channeling.
To assess the effect of equivariance, our backbone is made of 10 convolution layers with kernel size equal to 2, ELU non-linearity and  Max Pooling downsampling every two layers with kernel size equal to 2. Different degrees of equivariance were tested, namely, Classical CNN "translation equivariant", Equivariant 90° ($N$=$4$), Equivariant 45° ($N$=$8$), Equivariant 18° ($N$=$20$), Equivariant 10° ($N$=$36$), and finally the Equivariant SO(2). Equivariant models are generated based on \textit{e2cnn} implementation \cite{gen-E2-eq}. 
We trained the model on one sequence only, without any data augmentation and for 100 epochs. Number of parameters, optimizer, learning rate and random seeds were fixed.

\begin{figure}[t]
\begin{center}
\includegraphics[width=\linewidth, trim=0mm 1mm 0mm 0mm, clip]{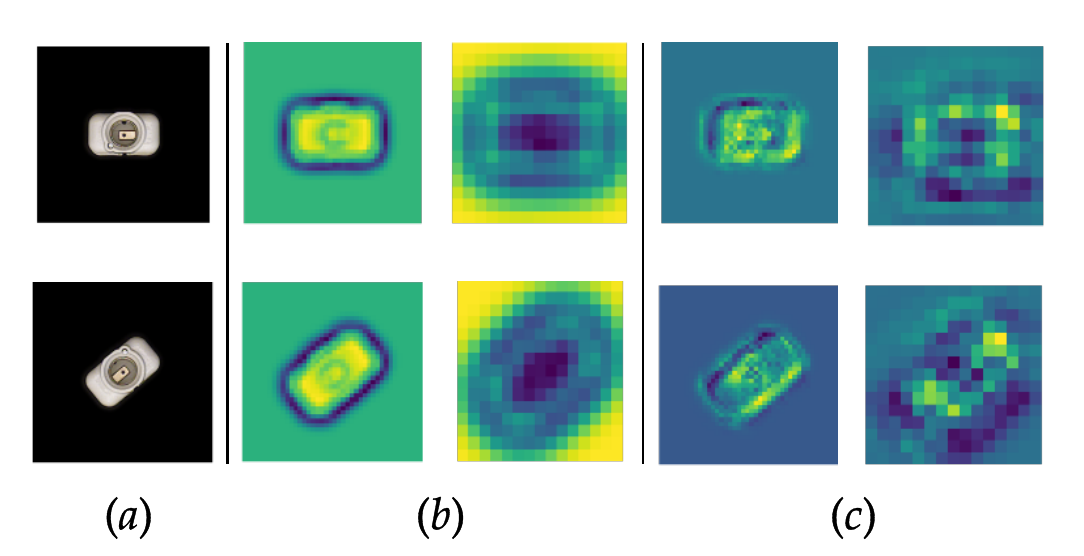}
\end{center}
\caption{ \textbf{Extracted feature maps -} difference between: (b) equivariant CNN and (c) classical CNN. The used samples (a) are from the T-LESS Dataset~\cite{T-LESS}.}
\label{fig:tless-feat}
\end{figure}

\begin{figure}[t]
\begin{center}
\includegraphics[width=\linewidth, trim=0mm 5mm 0mm 0mm, clip]{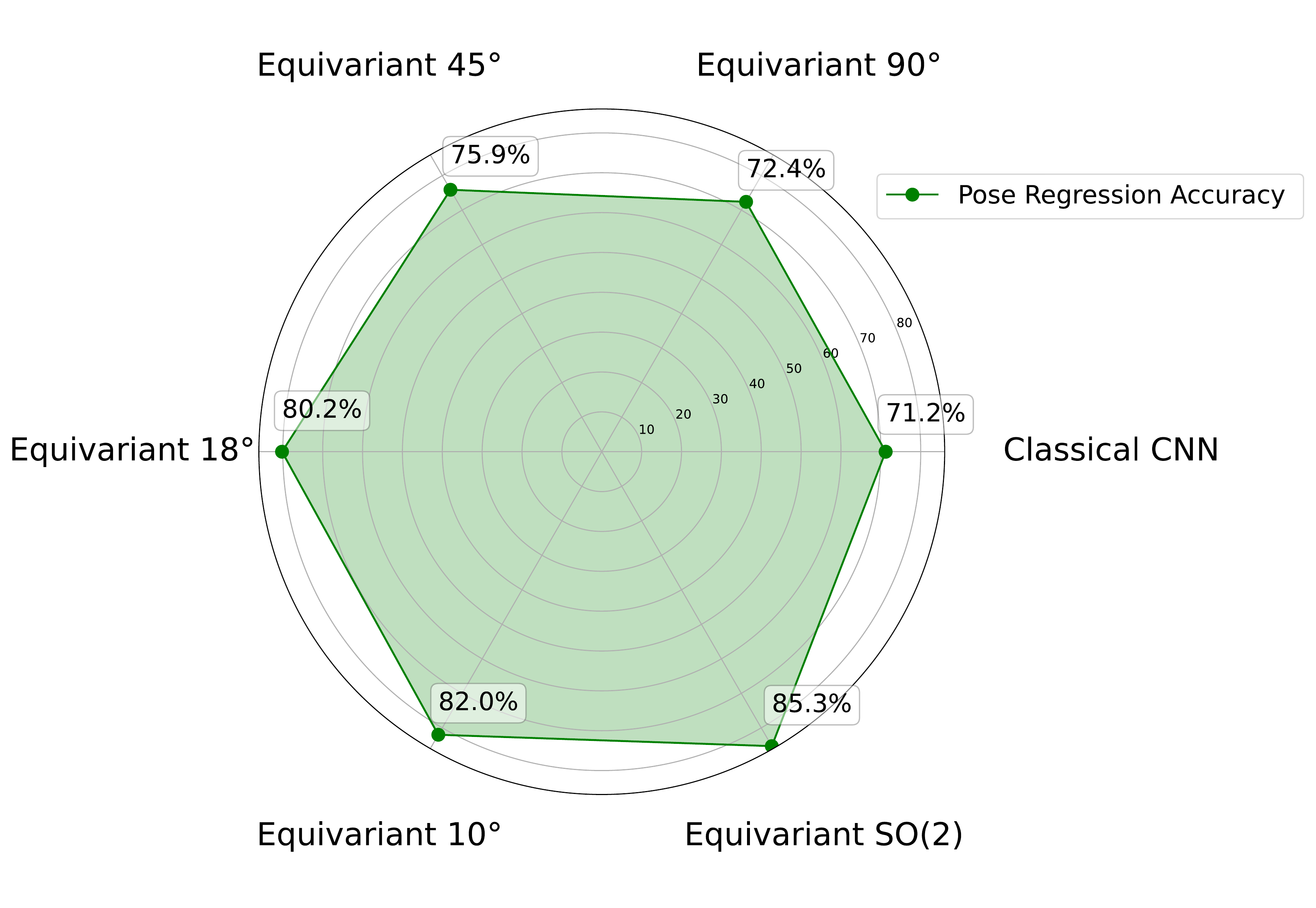}
\end{center}
\caption{
 \textbf{Equivariant models comparison -} On sequence of `object 5' from the T-less train dataset~\cite{T-LESS}, we can see the increase in the model accuracy when increasing the equivariant group.
}
\label{fig:tless-acc}
\end{figure}

\noindent Figure \ref{fig:tless-acc} reports the proportion of samples for which the predicted pose error is below 10cm, 10°. We observed that increasing the level of equivariance, \textit{i.e.,} decreasing the discrete sampling angle, leads to increasing the performance of the pose estimation model. Furthermore, the best reported performance has been achieved by the $\mbox{SO}(2)$ continuous rotation equivariance. The metric used here does not follow the standard T-LESS metric since it is only used for model variants comparison.\\
\noindent The difference between rotation-equivariant and classical CNN features is highlighted in Figure~\ref{fig:tless-feat}. By using images with different orientations (Figure \ref{fig:tless-feat}$(a)$) as input, the same transformation links extracted feature maps from different stages of the model $(b)$. In contrast, this is not the case with the classical CNN where the obtained feature maps are not rotated versions of each other $(c)$. 

\vspace{2mm}
\noindent\textbf{Datasets.}

\begin{table*}[t]
    \centering
    \begin{tabular}{l|c c c c}
    \hline
 & \textit{King's College} & \textit{Old Hospital} & \textit{Shop Facade} & \textit{St. Mary} \\
\hline
DenseVLAD + Inter. (Baseline)\cite{torii201524} & 1.48/4.45 & 2.68/4.63 & 0.90/4.32 &  1.62/6.06   \\
PoseNet (PN)\cite{PoseNet}            & 1.92/5.40 & 2.31/5.38 & 1.46/8.08 & 2.65/8.48 \\
PN learned weights~\cite{apr-cvpr-2017}         & 0.99/\textbf{1.06} & 2.17/2.94 & 1.05/3.97 & 1.49/3.43   \\
BayesianPN~\cite{kendall2016modelling}     & 1.74/4.06 & 2.57/5.14 & 1.25/7.54 & 2.11/8.38  \\
LSTM-PN~\cite{walch2017image}    & 0.99/3.65 & 1.51/4.29  & 1.18/7.44 & 1.52/6.68   \\
SVS-Pose ~\cite{naseer2017deep}   & 1.06/2.81 & 1.50/4.03 & 0.63/5.73 & 2.11/8.11   \\
GPoseNet~\cite{cai2019hybrid}     & 1.61/2.29 & 2.62/3.89 & 1.14/5.73 & 2.93/6.46   \\
MapNet\cite{brahmbhatt2018geometry}    & 1.07/1.89 & 1.94/3.91 & 1.49/4.22 & 2.00/4.53  \\
IRPNet~\cite{shavit2021we}     & 1.18/2.19 & 1.87/3.38 & 0.72/3.47 & 1.87/4.94   \\
\hline
MS-Transformer \cite{Shavit_2021_ICCV} & 0.83/1.47 & 1.81/\textbf{2.39} & 0.86/3.07 & 1.62/3.99 \\
TransPoseNet \cite{abs-2103-11477} & \textbf{0.60}/2.43 & 1.45/3.08 & \textbf{0.55}/3.49 & 1.09/4.99 \\
\hline
\textit{E-PoseNet} (Ours) & 0.95/1.63 & \textbf{1.43}/2.64 & 0.60/\textbf{2.78} & \textbf{1.00}/\textbf{3.16} 
\\ \hline
\end{tabular}
\caption{\textbf{Comparative analysis of pose regressors on Cambridge Landmarks dataset (outdoor localization)~\cite{PoseNet} -} We report the median position/orientation error in
meters/degrees for each method. Best results are highlighted in bold.}
\label{tab:cambridge_res}
\end{table*}

\begin{table*}[tbh]
\centering
    \begin{tabular}{l|c c c c c c c}
\hline
 & \textit{Chess} & \textit{Fire} & \textit{Heads} & \textit{Office} & \textit{Pumpkin} & \textit{Kitchen} & \textit{Stairs} \\ \hline
DenseVLAD + Inter.\cite{torii201524} & 0.18/10.0
& 0.33/12.4 & 0.15/14.3 & 0.25/10.1 & 0.26/9.42 & 0.27/11.1 & \textbf{0.24}%
/14.7 \\
PoseNet (PN)~\cite{PoseNet} & 0.32/8.12 & 0.47/14.4
& 0.29/12.0 & 0.48/7.68 & 0.47/8.42 & 0.59/8.64 & 0.47/13.8 \\
PN learned weights~\cite{apr-cvpr-2017} & 0.14/4.50
& 0.27/11.8 & 0.18/12.1 & 0.20/5.77 & 0.25/4.82 & 0.24/5.52 & 0.37/10.6 \\
BayesianPN~\cite{kendall2016modelling} & 0.37/7.24 &
0.43/13.7 & 0.31/12.0 & 0.48/8.04 & 0.61/7.08 & 0.58/7.54 & 0.48/13.1 \\
LSTM-PN~\cite{walch2017image} & 0.24/5.77 & 0.34/11.9 &
0.21/13.7 & 0.30/8.08 & 0.33/7.0 & 0.37/8.83 & 0.40/13.7 \\
GPoseNet~\cite{cai2019hybrid} & 0.20/7.11 & 0.38/12.3 &
0.21/13.8 & 0.28/8.83 & 0.37/6.94 & 0.35/8.15 & 0.37/12.5 \\
GeoPoseNet~\cite{apr-cvpr-2017} & 0.13/4.48 &
0.27/11.3 & 0.17/13.0 & 0.19/5.55 & 0.26/4.75 & 0.23/5.35 & 0.35/12.4 \\
MapNet~\cite{brahmbhatt2018geometry} & \textbf{0.08}/3.25
& 0.27/11.7 & 0.18/13.3 & 0.17/\textbf{5.15} & 0.22/4.02 & 0.23/%
\textbf{4.93} & 0.30/12.1 \\
IRPNet~\cite{shavit2021we} & 0.13/5.64 & 0.25/%
9.67 & 0.15/13.1 & 0.24/6.33 & 0.22/5.78 & 0.30/7.29 & 0.34/11.6 \\
AttLoc~\cite{wang2019atloc} & 0.10/4.07 & 0.25/11.4 &
0.16/11.8 & 0.17/5.34 & 0.21/4.37 & 0.23/5.42 & 0.26/10.5
\\
\hline
MS-Transformer \cite{Shavit_2021_ICCV} & 0.11/4.66 & 0.24/\textbf{9.60} & 0.14/12.2 & 0.17/5.66 & 0.18/4.44 & \textbf{0.17}/5.94 & 0.26/8.45 \\
TransPoseNet \cite{abs-2103-11477} & \textbf{0.08}/5.68 & 0.24/10.6 & \textbf{0.13}/12.7 & 0.17/6.34 & 0.17/5.60 & 0.19/6.75 & 0.30/\textbf{7.02} \\
\hline
\textit{E-PoseNet} (Ours) &\textbf{0.08/2.57} & \textbf{0.21}/11.0  & 0.16/\textbf{10.3}& \textbf{0.15}/6.80 & \textbf{0.16/3.82} & 0.20/6.81&  \textbf{0.24}/9.92 \\ \hline
\end{tabular}
\caption{\textbf{Comparative analysis of pose regressors on the 7-Scenes dataset
(indoor localization)~\cite{Shotton_2013_CVPR} -} We report the median position/orientation error in
meters/degrees for each method. Best results are highlighted in bold.}
\label{tab:7scenes_res}
\end{table*}

\begin{figure}[t]
\begin{center}
\includegraphics[width=\linewidth, trim=0mm 0mm 0mm 0mm, clip]{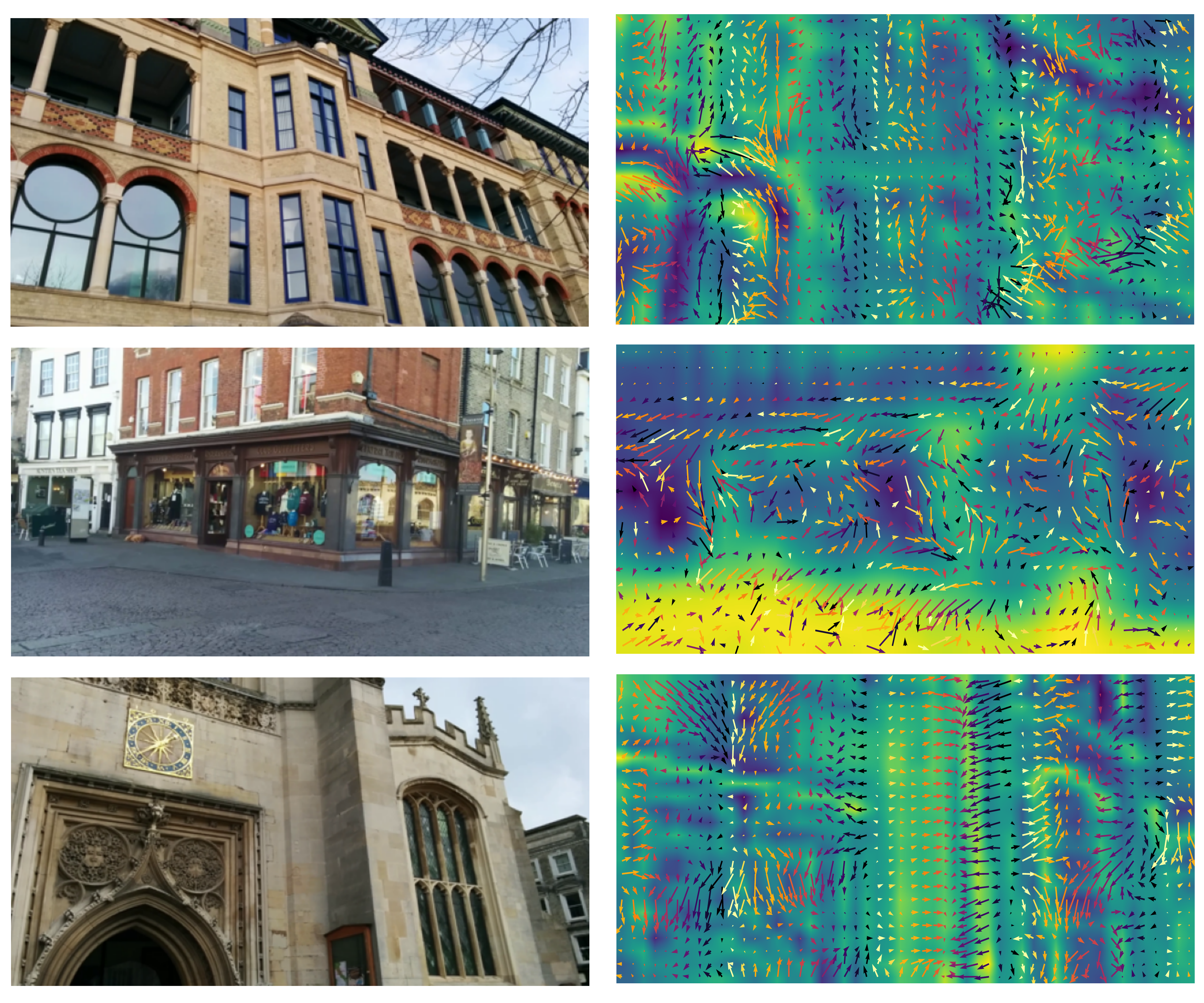}
\end{center}
\caption{
\textbf{Feature visualization for Cambridge Landmarks -}
Visualization of samples images from the Cambridge Landmarks dataset \cite{PoseNet} (left), along with their respective $\se$ group representations learned by \textit{E-PoseNet} (right).
}
\label{fig:ck_viz}
\end{figure}

\vspace{1mm}
\noindent\textbf{\textit{Cambridge Landmarks --}}
We use this dataset~\cite{PoseNet} to evaluate the performance of \textit{E-PoseNet} in outdoor camera relocalization. It is a large scale dataset taken around Cambridge University, containing original videos labelled with 6-DoF camera poses and a visual reconstruction of the scene (spatial extent of $\sim 900-5500 m^{2}$). We train and evaluate \textit{E-PoseNet} on four scenes (see Table \ref{tab:cambridge_res}). Furthermore, a few samples are used to visualize the obtained \textit{E-PoseNet} feature fields. Figure \ref{fig:ck_viz} shows that they directly support representations of $\se$ and are therefore enriched with some notion of orientation, visualized in a vector field form. On the contrary classical CNNs do not encode geometric information directly into their feature space.

\noindent\textbf{\textit{7-Scenes --}}
For indoor camera localization, we use the 7-Scenes dataset~\cite{Shotton_2013_CVPR} which is a collection of tracked RGB-D camera frames for indoor scenes with a spatial extent of $\sim 1-10 m^{2}$. Only RGB images are used in our experiments. 

\noindent Finally, the two datasets present various challenges, \textit{i.e.,} occlusion, reflections, motion blur, lighting conditions, repetitive textures, and variations in viewpoint and trajectory.

\vspace{2mm}
\noindent\textbf{Comparative Analysis of Camera Pose Regressors.}
We compare the performance of \textit{E-PoseNet} with state-of-the-art APR methods for camera localization in both outdoor and indoor scenes. First, we tested the performance on the Cambridge Landmarks dataset, for which we provide the median position and orientation errors in Table~\ref{tab:cambridge_res}. We also compare the performance of \textit{E-PoseNet} with respect to the state-of-the-art monocular pose regressors reporting on the 7-Scenes dataset. Table~\ref{tab:7scenes_res} contains the results. From the results on both datasets, and in comparison with APR methods, we conclude that the proposed \textit{E-PoseNet} achieves the lowest location error across all the outdoor and indoor scenes, and the lowest orientation error across the majority of them. It also competes with most recent transformer-based architectures \cite{Shavit_2021_ICCV,abs-2103-11477} on these datasets.

\vspace{1mm}
\noindent\textbf{Implementation Details.} 
We tested different architectures for the equivariant backbone, with ResNet18 being the most suitable model in our experiments, from both model size and  performance perspectives.
We trained our model for $5-10$k epochs using Adam optimizer~\cite{KingmaB14}, with $\beta_{1}=0.9, \beta_{2}=0.999$, $\epsilon=10^{-5}$ and a batch size of 256. During the training phase, we rescaled the image so that its smaller length is $256$ pixels followed by a random $224 \times 224$ crop. No further data augmentation was used.

\vspace{1mm}
\noindent\textbf{Limitations.}
While we focus on introducing equivariant operations for the feature extraction part of the APR pipeline, the following stages (\textit{i.e.} embedding, regression) do not have the same property, resulting in breaking the equivariance of the overall pipeline. Another limitation of the proposed APR model is the longer time required for equivariant CNN models as compared to classical CNN ones. Note that this is only during training, while the inference time is similar for both types of models.

\section{Conclusions}
\label{sec:conclusions}

This paper presents a new direction for the problem of camera pose regression leveraging equivariant features to encode more geometric information about the input image. By using an $\se$-equivariant feature extractor, our model is able to outperform existing methods on
both outdoor and indoor benchmarks.
Furthermore, we conclude that the equivariant properties of deep learning models that are used for geometric reasoning offer a promising direction for reaching the potential of absolute pose regression. 

{\small
\bibliographystyle{ieee_fullname}
\bibliography{main}
}

\end{document}